\documentclass[10pt, conference, compsocconf]{IEEEtran}
\pdfoutput=1
\usepackage{times}
\usepackage{epsfig}
\usepackage{graphicx}
\usepackage{amsmath}
\usepackage{amssymb}

\usepackage{algorithm}
\usepackage{algorithmicx}
\usepackage{algpseudocode}
\usepackage{amsmath}
\usepackage{multicol}
\usepackage{multirow}
\usepackage{booktabs}
\usepackage{threeparttable}
\usepackage{enumerate}   
\usepackage{lipsum}
\begin{document}
	\title{An End-to-end Video Text Detector with Online Tracking \vspace{-2em}}


	% \author{Anonymous ICDAR submission
	% \\Paper ID 99}
	
	\author{\IEEEauthorblockN{
			Hongyuan Yu\textsuperscript{13*$\dagger$},
			Chengquan Zhang\textsuperscript{2*},
			Xuan Li\textsuperscript{2},
			Junyu Han\textsuperscript{2},
			Errui Ding\textsuperscript{2}, and
			Liang Wang\textsuperscript{134}}
		\IEEEauthorblockA{
			\textsuperscript{1}University of Chinese Academy of Sciences (UCAS)\\
			\textsuperscript{2}Department of Computer Vision Technology(VIS), Baidu Inc.\\
			\textsuperscript{3} Center for Research on Intelligent Perception and Computing (CRIPAC), \\
			National Laboratory of Pattern Recognition (NLPR) \\
			%\textsuperscript{4}Center for Excellence in Brain Science and Intelligence Technology (CEBSIT), \\
			%Institute of Automation, Chinese Academy of Sciences (CASIA) \\
			\textsuperscript{4}Chinese Academy of Sciences Artificial Intelligence Research (CAS-AIR) \\
			Email: 
			hongyuan.yu@cripac.ia.ac.cn,
			\{zhangchengquan,lixuan12,hanjunyu,dingerrui\}@baidu.com,
			wangliang@nlpr.ia.ac.cn}
			\vspace{-2em}
	}

	
	%Telephone: (800) 555--1212, Fax: (888) 555--1212}


	\maketitle

    \newcommand\blfootnote[1]{% 
    \begingroup 
    \renewcommand\thefootnote{}\footnote{#1}% 
    \addtocounter{footnote}{-1}% 
    \endgroup 
    }
    
	\blfootnote{\textsuperscript{$\ast$}Equal contribution. \textsuperscript{$\dagger$}This work is done when Hongyuan Yu is intern at VIS, Baidu Inc.}
	
	%\vspace{-0.5cm} 
	%%%%%%%%% ABSTRACT
	\begin{abstract}
		%Text in videos usually acts as important semantic cues, which is helpful to video analysis. 
    Video text detection is considered as one of the most difficult tasks in document analysis due to the following two challenges: 1) the difficulties caused by video scenes, i.e., motion blur, illumination changes, and occlusion; 2) the properties of text including variants of fonts, languages, orientations, and shapes. Most existing methods attempt to enhance the performance of video text detection by cooperating with video text tracking, but treat these two tasks separately. In this work, we propose an end-to-end video text detection model with online tracking to address these two challenges. Specifically, in the detection branch, we adopt ConvLSTM to capture spatial structure information and motion memory. In the tracking branch, we convert the tracking problem to text instance association, and an appearance-geometry descriptor with memory mechanism is proposed to generate robust representation of text instances. By integrating these two branches into one trainable framework, they can promote each other and the computational cost is significantly reduced. Experiments on existing video text benchmarks including ICDAR2013 Video, Minetto and YVT demonstrate that the proposed method significantly outperforms state-of-the-art methods. Our method improves F-score by about 2\% on all datasets and it can run realtime with 24.36 fps on TITAN Xp.

	\end{abstract}

    \section{Introduction}
	With the rapid development of mobile internet, video-related applications become more and more popular in our daily life. The analysis of videos therefore becomes an important task for practical applications. Among various types of objects appearing in videos, text usually contains abundant semantic information and palys an important role in many applications, such as video annotation, multimedia retrieval and industrial automation~\cite{Ye2015TextDA,Yin2016TextDT}. 
	
	In the last few years, we have witnessed significant efforts in tackling video text detection. Previous works on this problem~\cite{Tian2016SceneTD,Wang2017EndtoEndST,Yang2017AUV,pei2018scene} are generally carried out in two steps: text in individual frame is detected first, the data association is then performed. However, those two-step methods suffer from the following problems: (1) single-frame detection in video does not make full use of the temporal context in the video; (2) tracking after detection needs additional networks to extract tracking features, which leads to additional computational cost, and most of these tracking methods are offline, which is limited in practical applications; (3) these two parts are separately trained, and they cannot fully utilize the supervision information of each other. Video text detection and tracking tasks are closely related。
	%Video text tracking can help video text detection. So it is not wise to isolate video text detection and tracking. 
	
	%fig_intro
	\begin{figure}[tbp]
		\centering
		\includegraphics[scale=0.45]{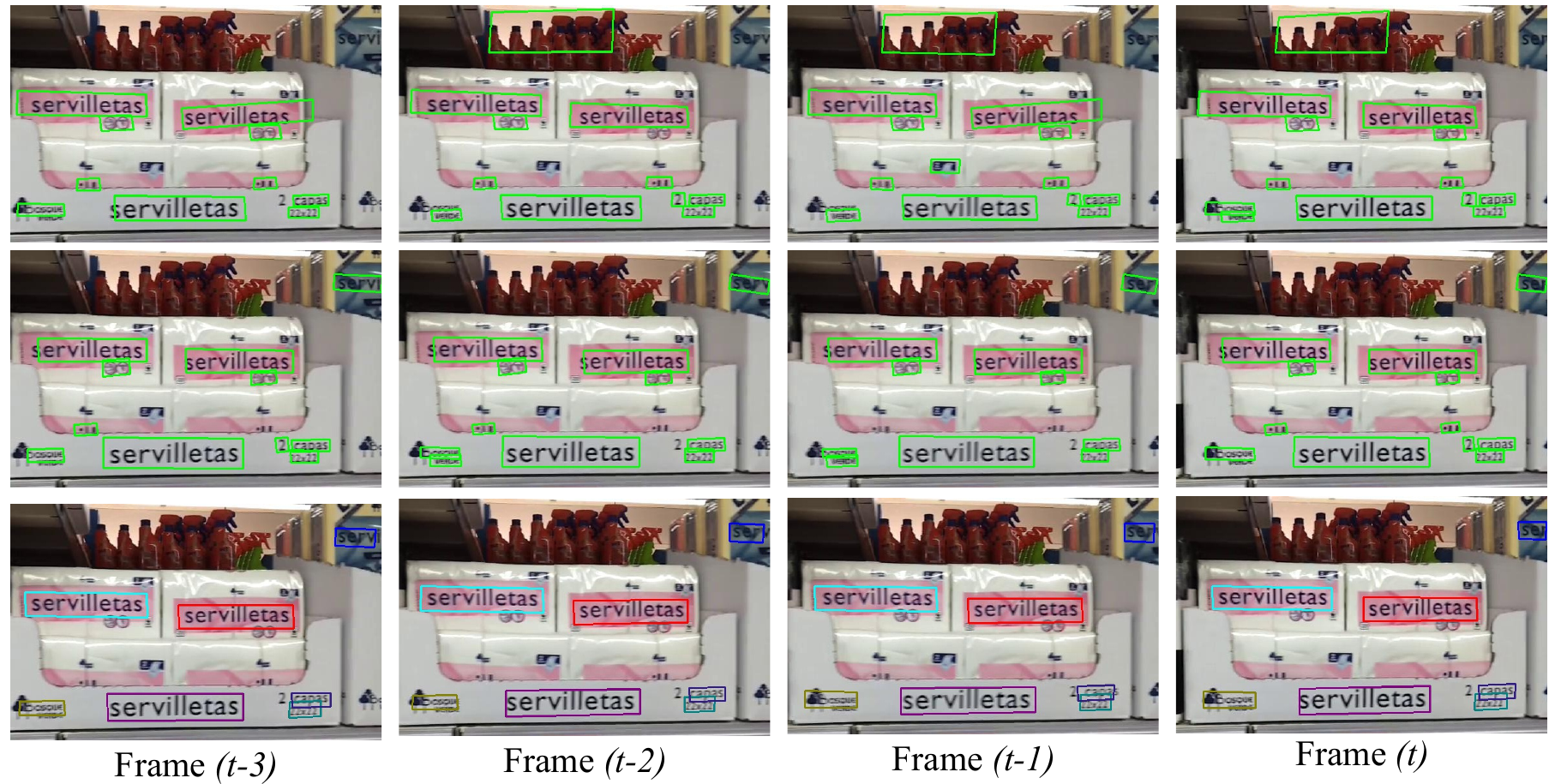}
		\caption{Video text detection and tracking. The first row is the detection results of EAST~\cite{zhou2017east}. The second row is the results from our video text detection branch with ConvLSTM. The final row is our detection results with online tracking, where boxes with the same color in different frames belong to the same trajectory. (Best view in color.)} 
		\label{fig:intro}
		\vspace{-0.75cm} 
	\end{figure}
	
	In this work, we present a novel end-to-end video text detector with online tracking. 
	%By tackling detection and tracking together, they can exploit the supervision information of each other. 
	By integrating detection and tracking together, they can exploit the supervision information of each other. The whole pipeline of our proposed approach is depicted in Fig.~\ref{fig:arch}. Specifically, to take full advantage of the temporal domain information and texture properties of scene text while preserving its structure, ConvLSTM~\cite{xingjian2015convolutional} layer is introduced to the video text detection branch. In addition, the appearance-geometry descriptor ($AGD$) and corresponding estimated appearance-geometry descriptor ($EAGD$) are proposed to model short-term target association for online tracking. Simultaneously, these descriptors are updated in time steps to capture the information of the long-term multiple targets changing process, $i.e.$, new target entry and old target departure. As shown in Fig.~\ref{fig:intro}, our method can significantly improve the performance. 
	
	The contributions of our method are summarized as follows: (1) to the best of our knowledge, this is the first end-to-end video text detection and online tracking framework; (2) we introduce ConvLSTM to our detection branch, which is very useful for capturing spatial-temporal information; (3) the proposed appearance-geometry descriptor has been proven to be robust and effective for multiple text instance association; (4) extensive experiments have shown the effectiveness of our method, and we have obtained state-of-the-art results in multiple public benchmarks.

	%---------------------------------------------------------------------------------------------------
	\section{Related Work}\label{sec:related_work}
	Single frame text detection has made great progress recently. However, for video text detection, how to make full use of the context information in video is still not addressed well. In this section, the development of single image detection and video text detection will be reviewed.
	
	\textbf{Single frame text detection} 
	It is known that numerous methods for text detection have been successfully proposed in recent years. 
	%Video text detection can be divided into the following two categories: single frame detection and video mode detection. 
	Specifically, component based methods, segmentation based methods and detection based methods are the main kinds of single frame detection. \emph{Component based methods}~\cite{epshtein2010detecting,shi2017detecting} usually detect parts or components of text first, after then a set of complex procedures including component grouping, filtering and word partition are followed to obtain final detection results in word-level. %Component based methods are flexible to solve arbitrarily oriented text, but neither robust nor end-to-end trainable. 
	\emph{Segmentation based methods}~\cite{zhang2016multi,wu2017self} regard all the pixels in one word / text-line as a whole instance. %Similar to instance-level semantic segmentation methods, FCN~\cite{long2015fully} is the most commonly used framework. Segmentation based methods 
	It can handle arbitrary shape of text, but rely too much on fine grained segmentation and also need some discontinuous post-processing operations. \emph{Detection based methods}~\cite{he2017deep,zhou2017east} draw inspiration from general object detection, and output text detection results in word / text-line level directly. 
	
	\textbf{Video text detection} 
	Standing on the shoulder of single frame text detection, many video mode text detection methods have also been proposed. 
	Researchers try to enhance the detection result through tracking, namely tracking based text detection methods. These methods utilize some specific tracking techniques, such as multi-strategy tracking methods~\cite{zuo2015multi}, dynamic programming~\cite{Tian2016SceneTD} and network flow based methods~\cite{Yang2017AUV}, to track text and then heuristically combine detected results in passed frames. But they are essentially based on single-frame detection methods, and the training of detector is independent of the tracking. Moreover, they do not make full use of the abundant temporal information of video. Wang et al.~\cite{Wang2018SceneTD} notice that the cues of background regions can promote video text detection. However, they only consider the short-term dependencies. There is no doubt that some structures like optical flow, Conv3D~\cite{ji20133d} and ConvLSTM~\cite{xingjian2015convolutional} are efficient to catch spatial-temporal information, which have been explored in general object tracking. Inspired by that, in this paper, we employ ConvLSTM in video text detection branch in our framework. With the help of long-term spatial-temporal memory and online tracking, our end-to-end video text detector achieves better performance.

	\section{Approach}\label{sec:approach}
	\begin{figure*}
		\centering
		\includegraphics[scale=0.65]{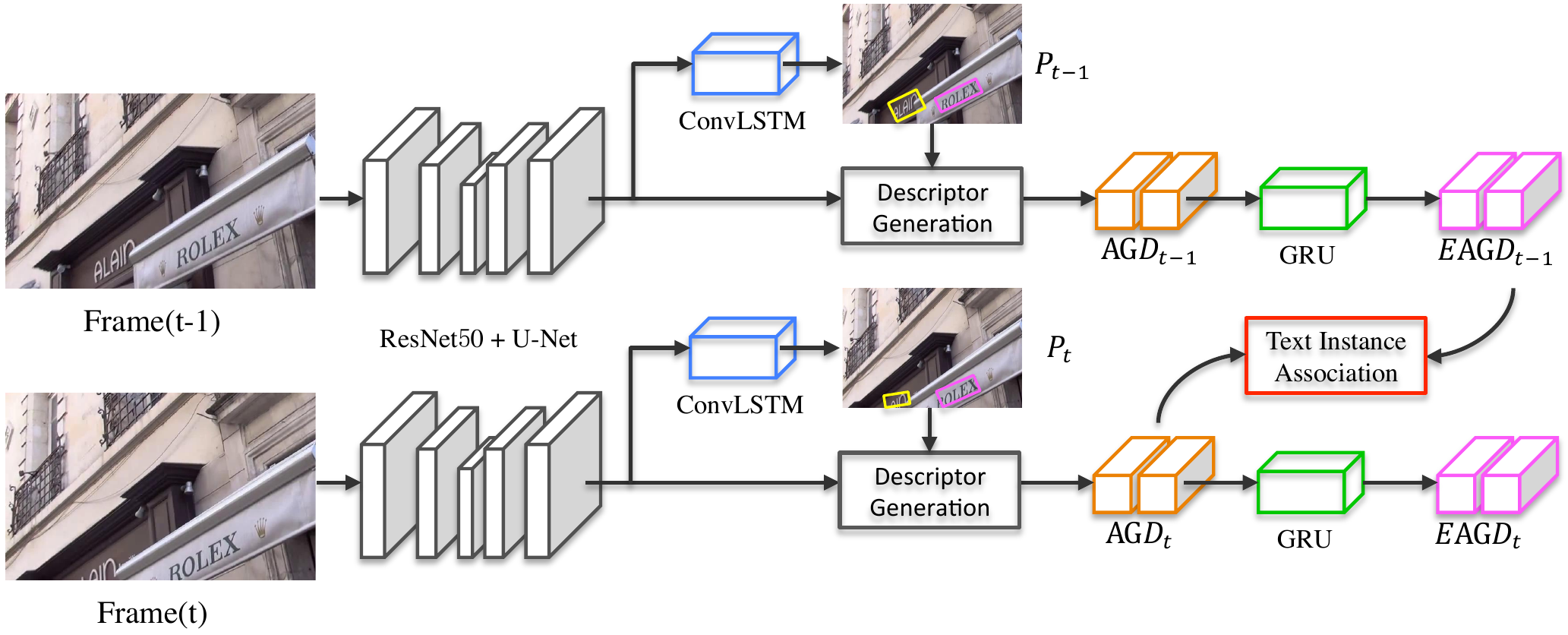}
		\caption{The proposed architecture of video text detection with online tracking.}
		\label{fig:arch}
		\vspace{-0.65cm} 
	\end{figure*}
	
	% In this section, we describe the proposed approach in details. Firstly, the overall architecture shown in Fig.~\ref{fig:arch} will be introduced briefly. Then, the main modules including video text detection and tracking are explained. Finally, an efficient and robust inference procedure for online trajectory generation is presented.
	%Finally, an efficient and robust inference procedure for online text tracking is presented.
	%Finally, an efficient online trajectory generation method based are presented.
	%--------------------------------
	\subsection{Overall Architecture}
	%Table \ref{tab:compare_still&video} shows the benefits from different parts compare with baseline model.
	The proposed method, as shown in Fig.~\ref{fig:arch}, integrates video text detection and tracking in an unified framework through the descriptor generation module. Given a video, all frames should pass a backbone network (ResNet50~\cite{he2016deep} + U-Net~\cite{ronneberger2015u}) to extract common features for detection and tracking.
	%The channel number of common features is 128 and its size is 1/4 of the original image. 
	For video text detection, we adopt the anchor-free regression manner~\cite{zhou2017east} %used in EAST~\cite{zhou2017east}
	to detect the quadrangles of words in a per-pixel manner. 
	Notice that a ConvLSTM block is followed with common features to extract spatial-temporal information. The benefits from ConvLSTM block are shown in Tab.~\ref{tab:detection_ic13}. 
	%It should be noted that a ConvLSTM block is followed with common features to extract spatial-temporal information.
	% The benefits from ConvLSTM block are shown in Tab.~\ref{tab:detection_ic13}. 
	For video text tracking, detected proposals and common feature maps of current frame are fed into the descriptor generation module first, and then corresponding appearance-geometry descriptors $AGD_t$ are output. 
	%To be specific, we keep top $K=10$ detected text instance proposals of each frame and feed them into descriptor generation module in training phase. 
	%The descriptor generation module is described thoroughly in Sec.~\ref{sec:ttb}. 
	%which uses ROI transform layer~\cite{sun2018textnet} to extract appearance features, and encodes geometry features by embedding the coordinate values of detected bounding boxes. 
	In order to associate text instances in sequential frames,  the descriptors of frame ($t-1$) namely $AGD_{t-1}$ are passed through GRU units to generate $EAGD_{t-1}$, meaning the estimated states of appearance-geometry descriptors at time $t$. 
	%$D_t$ and $ED_{t}$ can also be extracted from frame($t$) in the same way. 
	After then, a similarity matrix is build on $EAGD_{t-1}$ and $AGD_t$, where two text proposals belonging to the same trajectory should have a small metric distance value in the similarity matrix. With the help of online text tracking, our methods are able to improve the performance of video text detection.
	
	\subsection{Text Detection Branch} \label{sec:tdb}
	Text in videos always appears in sequential frames with abundant temporal information. In most of existing methods, video text detection is usually performed in individual frame or integrated temporal information with short term dependencies. 
	%But none of them are efficient or effective enough to capture temporal information. 
	To address this problem, we incorporate a ConvLSTM block into our text detection branch to propagate frame-level information across time while maintaining the structural properties. Accordingly, the formulation of inferring feature maps $F_{t}$ at the $t$-th frame is:
	\begin{equation}
	\setlength{\abovedisplayskip}{5pt}
	\setlength{\belowdisplayskip}{5pt}
	(F_{t}, s_{t}) = ConvLSTM(M(I_{t}), s_{t-1})
	\end{equation}
	In this equation, $M(I_{t})$ is the common feature maps of the $t$-th frame($I_{t}$) obtained by backbone network. $s_{t-1}$ and $s_{t}$ mean the hidden sates of ConvLSTM at time $t-1$ and time $t$ respectively. By this way, features can be directly modulated by their previous frames and recursively depend on other frames in a long time range.
	
	After integrating temporal information, convolution operation is applied to make dense per-pixel predictions in word-level. Similar to EAST~\cite{zhou2017east}, pixels within the quadrangle annotation of text instance
	%the range of text instances, which are generally labeled as quadrangles, 
	are considered as positive. For each positive sample, the offsets to the 4 vertexes of the quadrangle are predicted at the following 8 channels. Therefore, the loss of detection branch is composed of two terms: text/non-text classification term and quadrangle offset regression term. The detailed definition of detection loss in $t$-th frame is illustrated as follows:
	\begin{equation}
	\setlength{\abovedisplayskip}{5pt}
	\setlength{\belowdisplayskip}{5pt}
	L_{det}(t) = L_{cls}(t) + \alpha L_{off}(t)
	\label{equa:loss}
	\end{equation}
	where $L_{cls}(t)$ measures the 
	%difference between the predicted and ground-truth score maps by cross-entropy loss ?lx
	text/non-text classification by dice loss 
	and $L_{off}(t)$ is the smooth-L1 loss to measure the quality of regression  offset. $\alpha$ is a hyper-parameter, %to control the trade-off between these two terms, 
	which is set to 5 in our experiments. In addition, we use NMS (Non-maximum suppression) to get preliminary detection results and feed top-K proposals into the next tracking branch.
	
	%the association branch through descriptor generation module, which is presented in the next section.
	
	%----------------------------------------------------------
	\subsection{Text Tracking Branch}
	\label{sec:ttb}
	%When the text detection described in the preceding section is completed, we need to associate text candidates in continuous frames. 
	%As we know, it is critical to represent and measure text instances during multiple text instance tracking. 
	%Most of general object tracking methods are based on the prediction of the bounding box offsets of text candidates in neighbouring frames, which is easy to fail 
	In order to improve the robustness of text instance representation in some difficult circumstances such as occlusion, motion blur, and etc, this section proposes an effective and efficient descriptor generation module to generate the descriptors of text candidates, which contains not only geometry features but also appearance features. And the descriptor of the same text candidate appears in the next frame is estimated through a GRU unit which can make use of the trajectory history information and capture the information of long-term multiple targets changing process. 
	%Finally, we convert the text instance association to pairwise matching problem in Sec.~\ref{sec:ae}.
	
	%\paragraph{Descriptor Generation}
	We define a novel descriptor, namely~\emph{appearance-geometry descriptor} ($AGD$), for each text candidate. The descriptors of the kept $K$ proposals from detection branch, denoted as $AGD_t$, contain two parts: the first part is the~\emph{appearance feature}, which is extracted by ROI Transform layer~\cite{sun2018textnet} from the valid regions of text candidates in the common feature maps $M(I_{t})$. And the second part is the~\emph{geometry feature} that is composed of the embedding values of quadrilateral coordinates. 
	%Fig.~\ref{fig:descriptor} shows how our descriptors are generated.
	%our appearance and geometry descriptor(D) contains two parts. The first part is the appearance feature, which is the feature in the common feature map cropped by text bounding box. And the second is the geometry feature. 
	%The generation of our tracking descriptor is presented in figure \ref{fig:descriptor}. 
	\begin{figure}[tbp]
		\centering
		\includegraphics[scale=0.6]{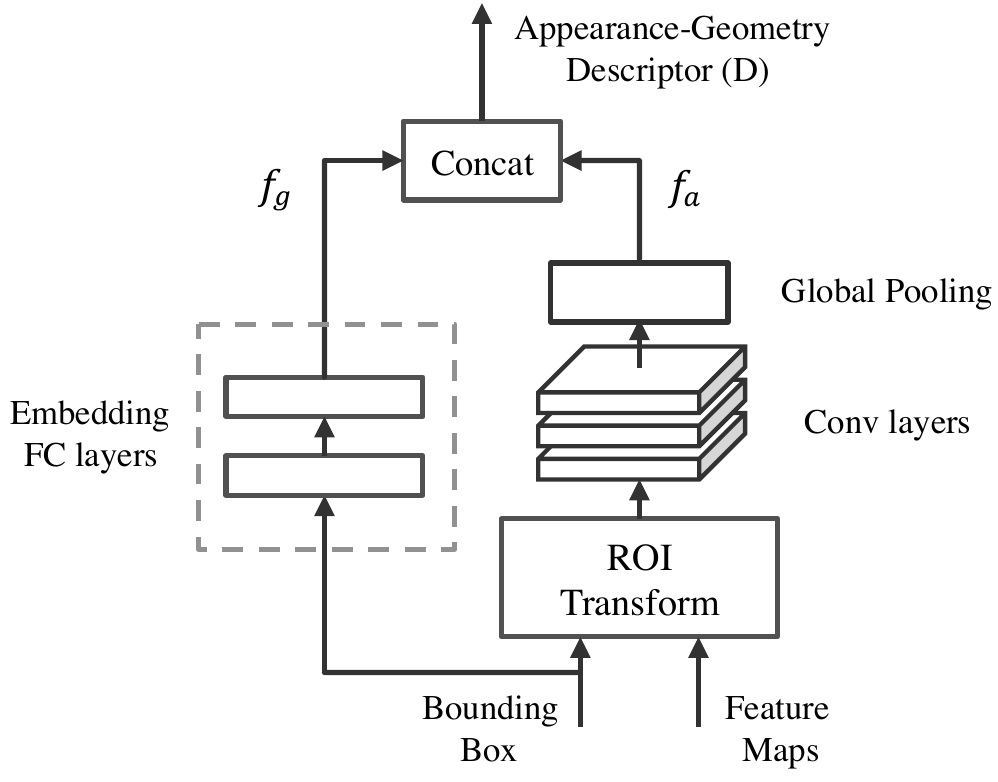}
		\caption{Descriptor Generation.}
		\label{fig:descriptor}
		\vspace{-0.65cm} 
	\end{figure}
	
	As shown in Fig.~\ref{fig:descriptor}, we firstly use ROI Transform layer to extract initial text feature block of $K$ text candidates from the shared feature map. %, whose shape is $K*128*8*256$. 
	Then three convolution layers with $3*3$ kernel and one global pooling layer are followed to generate the final appearance feature, denoted as $f^{a}_{t}$ for time $t$. 
	%, and the shape of $f^{a}_{t}$ is $K*128$. 
	Next, we feed all normalized coordinate vectors $\{g_{n}|n=0,...,7\}$ of detected $K$ proposals into the geometry embedding layers that are composed of two fully connection layers 
	%with $8$ hidden units 
	to get final geometry feature at time $t$, namely $f^{g}_{t}$. Finally, the appearance features $f^{a}_{t}$ and geometry features $f^{g}_{t}$ are concatenated to generate the appearance-geometry descriptor $AGD_{t}$, which can be formulated as follows:
	%Then we concatenate these two features' channel by $C$, The final descriptor $D(t)$ is:
	\begin{equation}
	AGD_{t} = Concat(\left[f^a_t, f^g_t\right])
	%D(t) = Conv_{1x1}(C(f^{a}_{t}, f^{g}_{t}))
	\label{equa:det}
	\end{equation}
	%Therefore, the dimension of descriptors $AGD_{t}$ is $K*136$.
	
	Then we feed the descriptor into a GRU unit to estimate the descriptor of the same instance appeared in the next frame, namely~\emph{estimated appearance-geometry descriptor} ($EAGD$). As we know, GRU is an efficient structure to capture the temporal changing information. Therefore, instead of building similarity matrix on appearance-geometry descriptors between two adjacent frames, we choose to match descriptors of the current frame with the estimated descriptors of the previous frame. The estimated descriptor $EAGD_{t}$ in current frame can be expressed as:
	\begin{equation}
	(EAGD_t, h_t) = GRU(AGD_t, mask_t * h_{t-1})  
	\end{equation}
	where $AGD_t$ is the appearance-geometry feature of text candidates obtained in current frame, $h_{t-1}$ is the hidden state value of GRU in previous frame. Specially, $mask_{t}$ is the hidden state mask to control whether we need to reset GRU hidden states or not. It will be set to zero when the instance does not exist in the previous frame; otherwise, it will be set to one.

	Video text tracking tries to match text instances belonging to the same object in adjacent frames while maintaining the identities of text instances.
	%Inspired from the siamese network~\cite{tao2016siamese} for matching pairwise object, 
	To simplify this problem, we convert the text instance association to pairwise matching by defining an association objective function, where descriptor representations should be close for the positive pairs and far for the negative pairs. Therefore, the contrastive loss is suitable for this task, then our tracking loss $L_{track}$ at time $t$ can be represented by:
	\begin{equation}
	%L_{track}(t) = \sum_{i=1}^{K}\sum_{j=1}^{k} Y^{i,j}(t) * S^{i,j}_{t} + (1-Y^{i,j}(t))*max(m - S^{i,j}_{t}, 0)^2
	%L_{track}(t) = \sum_{i=1}^{K}\sum_{j=1}^{K} yd^{2} + (1-y)max(margin-d,0)^2
	L_{track}(t) = \frac{1}{K^2} \sum_{i=1}^{K}\sum_{j=1}^{K} yd^{2} + (1-y)max(m-d,0)^2
	%L_{track} = \frac{1}_{2N} \sum_{n=1}^{N}yd^{2} + (1-y)max(margin-d,0)^2
	%L_{track} = \frac{mask^{p}}{2N} \sum_{n=1}^{N}yd^{2} + (1-y)max(margin-d,0)^2
	\end{equation}
	where $d$ denotes euclidean distance of text instances between adjacent frames and $y$ is the pairs label $L^{t}_{i,j}$ whose value is $1$ for positive pairs and $0$ for negative pairs. And $m$ is the margin value that is set to $1.0$ in our experiments. %??? same formulation
	%$margin$ is set to 1 in our experiments.
	
	Finally, combined with detection loss $L_{det}(t)$ in Eq.~\ref{equa:loss}, the full multi-task loss function is:
	%Finally the detection and tracking loss $L_{d\&t} $ is:
	\begin{equation}
	L_{d\&t} = \frac{1}{N} \sum_{t=1}^{N} L_{det}(t) + \beta L_{track}(t)
	\end{equation}
	where $N$ is length of video frames, and $\beta$ is a hyperparameter to control the trade-off between detection and tracking loss. $\beta$ is set to $0.1$ in our experiments, where different $\beta$ values have little effect on the final result. 
	
	\subsection{Inference}
	In the inference phase, we propose an efficient and robust online trajectory generation method to improve the performance of video text detection, which is present in algorithm~\ref{alg:track}. Besides, the inference speed of our method on TITAN Xp can reach 24.36fps.
	
	\floatname{algorithm}{Algorithm}
	\renewcommand{\algorithmicrequire}{\textbf{Input:}}
	\renewcommand{\algorithmicensure}{\textbf{Output:}}
	\begin{algorithm}
		\caption{Online Trajectory Generation}
		%\caption{Online text detection and tracking}
		\begin{algorithmic}[1]
			\Require A frame of current time step $I_{t}$, the previous detection results $D_{t-1}$, corresponding estimated appearance-geometry descriptors ($EAGD_{t-1}$), and tracklet set $T_{t-1}$ at time $t-1$.
			\Ensure Current frame detection results $D_{t}$,  the corresponding estimated appearance-geometry descriptors ($EAGD_{t}$) for next time step, and the tracklet set $T_{t}$.
			\State Feed the $I_{t}$ into the network to get the primary detection results $D^{*}_{t}$ by $\theta_{l}$ at time $t$, and obtain the corresponding appearance-geometry descriptors $AGD_{t}$.
			\State Calculate the similarity matrix $S_{t} \leftarrow similarity($ $EAGD_{t-1}, AGD_{t})$.
			\State Use Kuhn-Munkres algorithm with threshold value $\theta_{m}$ to find the matching pairs $M$.
			\State Update the part of tracklet set which find matching text instances in current frame, namely $T_{update}$
			%\State Mark the remaining tracklet set as $T_{end}$, where  $T_{end} \leftarrow T_{t-1} - T_{update}$
			\State Reward the matched instance confidence score by  $\tau * ln(length(tracklet))$. If the scores of no-matching candidates are higher than $\theta_{h}$, new trajectories $T_{new}$ are built for them.
			\State Obtain the full tracklet set of current time: $T_{t} \leftarrow T_{update} + T_{new}$
			\State Mark the corresponded detected boxes of $T_{t}$ at current time step as $D_{t}$, and feed the corresponding $AGD_{t}$ into the GRU to get the $EAGD_{t}$
		\end{algorithmic}
		\label{alg:track}
		%\vspace{-0.25cm} 
	\end{algorithm}
	\vspace{-2em}
	\section{Experiments}
	% In this section, we conduct our algorithm with ablation study
	%ablation study
	% to demonstrate its effectiveness. Firstly, we introduce three public datasets. %The detection and tracking evaluation metrics are described respectively in the second part.
	% Then, experimental configuration is presented. Finally, to show the effectiveness of our model, we compare our end to end model with state-of-the-art methods.
	%[width=\textwidth]
	%fig_res
	%\begin{figure*}
	%	\centering
	%	\includegraphics[scale=0.82]{figs/fig7.pdf}
	%	\caption{Some results of continuous frames on ICDAR 2013 (1st and 2nd rows), Minetto (3rd and 4th rows) and YVT (5th and 6th rows). (Best view in color.)}
	%	\label{fig:res}
	%\end{figure*}
	
	\subsection{Datasets}
	\begin{itemize}
		\item \textbf{ICDAR 2013 Video}~\cite{karatzas2013icdar}
		This dataset consists of 28 videos lasting from 10 seconds to 1 minute in indoors or outdoors scenarios. %, is a multilingual set. %was collected with several different languages. 
		13 videos used for training and 15 for testing. %testing videos taken by 4 different cameras.   %To cover a variety of possible hardware, 
		Its rame size ranges from 720 $ \mathrm{x} $ 480 to 1280 $ \mathrm{x} $ 960.
		%Its frame rate ranges from 24 to 30 fps and the frame size ranges from 720 $ \mathrm{x} $ 480 to 1280 $ \mathrm{x} $ 960.
		%The training set contains 13 videos and the test set contains 15 videos.
		
		%\item \textbf{ICDAR 2015 Video}~\cite{karatzas2015icdar}
		%This dataset, containing a training set of 25 videos %(13,450 frames) 
		%and a test set of 24 videos, %(14,374 )
		%was expanded by %expanded based on ？？？
		%the ICDAR 2013 Video. Since the ground truth of the test set is not available, we only use its training set. 
		
		\item \textbf{Minetto Dataset}~\cite{minetto2011snoopertrack}
		Minetto Dataset consists of 5 videos in outdoor scenes. The frame size
		is 640 $ \mathrm{x} $ 480 and all videos are used for test.% and the frame rate is 30 fps. 
		
		\item \textbf{YVT}~\cite{nguyen2014video}
		This dataset contains 30 videos, 15 for training and 15 for testing. Different from the above 2 datasets, it contains web videos except for scene videos.
		The frame size is 1280 $ \mathrm{x} $ 720. % and the frame rate ranges from 15 to 30 fps.
		
	\end{itemize}
		
	\subsection{Implementation Details}\label{sec:id}
	The ResNet50~\cite{he2016deep} pretrained on ImageNet is employed as our initialized model. Then Adam is employed to train our model with the initial learning rate being ${10}^{-4}$ which decays by 0.94 times every 10 thousand iterations. 
	%The cross-border protection is applied in the text tracking module to prevent crashing.
	All training videos are harvested from the training set of ICDAR2013 and YVT with data augmentation. Random crop and resize operations are applied for the first frame %in a video clip
	with the scale chosen from [0.5, 1.0, 2.0, 3.0]. Other frames in the video clip perform the same operation as the first frame. The frame interval is selected randomly from 1 to 5 to improve robustness. In our experiment, each video clip has 24 frames and every frame is resized and padded to 512 $ \mathrm{x} $ 512 during the training process. Each frame contains 10 detection boxes, including positive samples and negative samples. %These samples are distinguished by whether they can match the boxes in the previous and following frames.
	%The positive sample is a box whose preceding and following frames can match, while the negative sample is a box whose preceding and following frames do not match. 
	The shape of detection box extracted by ROI Transform layer is set to 8 $ \mathrm{x} $ 64. The size of the appearance descriptor and the geometry descriptor for one instance is 128 and 8, so the size of $AGD$ in our experiment is 136. All experiments are conducted on 8 P40 GPUs and each GPU has 1 batch.
	
	%\subsection{Ablation Study}n
	%\subsection{Evaluation of Exploiting Spatial-Temporal Information in Text Detection}
	
	\subsection{Evaluation of Video Text Detection}
	In this section, we evaluate the effect of the short-term and long-term memory for video text detection. As shown in Tab. \ref{tab:detection_ic13}, the optical flow is adopted and there is about 0.2\% drop in f-measure, 
	%we first use the optical flow to capture fast-changing information in video, and there is about 0.2\% drop in f-measure, 
	indicating that the common object detection tracking method is not applicable to video text. 
	Then, we exploit the ConvLSTM block in our detection branch and compare it with Conv3D~\cite{ji20133d}. 
	%, which detects 4 frames together in our experiment. 
	As can be seen in Tab. \ref{tab:detection_ic13}, both of these methods outperform single frame detection and the optical flow, and ConvLSTM gives a 0.66 points F-measure gain over Conv3D. 
	The improvement in video text detection is mainly due to the fact that 
	the temporal information in sequential frames is beneficial to video text detection since the text in video generally does not change as sharply as a general object. And as a valid long-term memory extractor, ConvLSTM can take advantage of more sequential frames information than Conv3D, resulting in an improvement in performance.
	%Then, we tentatively add conv3d, which detects 4 frames together in our experiment, to the network, and it is observed that the performance is much better. 
	
	%The improvement in detection performance is mainly due to the fact that the text in video generally does not change as sharply as a general object and Conv3D~\cite{ji20133d} can take advantage of more sequential frames information to retrieve some hard examples compared with the single frame detection and optical flow. Moreover, the f-measure is further improved by 0.66\% when the Conv3D is replaced with ConvLSTM, a more valid long-term memory extractor. Therefore, it is naturally believed that long-term memory will bring a significant improvement in the video detection.
	
	%Therefore, long-term memory will bring a significant performance improvement. Naturally, When we replaced conv3D with convLSTM, the f-measure was futher improve 0.66\% and it works in an online mode. 
	\renewcommand{\arraystretch}{1.5} %
	\begin{table}[tp]
		\centering
		\fontsize{9}{8}\selectfont
		\begin{threeparttable}
			\begin{tabular}{cccc}
				\toprule
				\multirow{1}{*}{Method}&
				Precision&Recall&F-measure\cr
				\midrule
				EAST~\cite{zhou2017east} &64.13&53.22&56.44 \cr
				Ours detection &75.08&52.28&61.64 \cr
				Ours with optical flow &68.49&55.69&61.43 \cr
				Ours with Conv3D &78.94&54.76&64.66 \cr
				Ours with ConvLSTM &79.97&55.21&65.32 \cr
				\bottomrule
			\end{tabular}
			\caption{\label{tab:detection_ic13}Comparison of video text detection performances on ICDAR 2013 dataset~\cite{karatzas2013icdar}.}
		\end{threeparttable}
		\vspace{-0.4cm} 
	\end{table}
	
	\renewcommand{\arraystretch}{1.5} %
	\begin{table}[tp]
		\centering
		\fontsize{9}{8}\selectfont
		\begin{threeparttable}
			\begin{tabular}{ccc}
				\toprule
				\multirow{1}{*}{Method}&MOTP&MOTA \cr
				\midrule
				Zuo et al. \cite{zuo2015multi} &73.07&56.37 \cr
                Pei et al. \cite{pei2018scene} &73.07&57.71 \cr
                %Our two stage &74.50 &61.92 \cr
                Ours with appearance descriptor &75.90&72.79 \cr
                Ours with geometry descriptor &76.66&74.04 \cr
                matching AGD with AGD &74.70&75.62 \cr
				matching AGD with EAGD &75.72&81.31 \cr
				\bottomrule
			\end{tabular}
			\caption{\label{tab:tracking}Comparison of video text tracking performances on Minetto dataset \cite{minetto2011snoopertrack}.}
		\end{threeparttable}
	\vspace{-0.65cm} 
	\end{table}
	
	\subsection{Evaluation of Video Text Tracking}
	Tab. \ref{tab:tracking} shows the influence of different types of tracking descriptors. We adopt the widely-used CLEAR MOT metrics~\cite{bernardin2008evaluating}, including MOTP (Multi-Object Tracking Precision) and MOTA (Multi-Object Tracking Accuracy) as text tracking evaluation metrics. The MOTP is the mean error of estimated positions for matched pairs of all frames, while the MOTA accounts for errors made by trackers, i.e., false positives, misses and mismatches.
	% anIt is the total error in estimated position for matchedobject-hypothesis   pairs   over   all   frames,   averagedby  the  total  number  of  matches  maded matching methods.% on the tracking indicator. %Overall, our method outperforms other methods on MOTA by a large margin. 
	At first, the appearance and geometry descriptors are studied respectively. Compared with appearance feature, the performance of using geometry features is 1.25\% ahead. However, there is a significant gain about 7\% when they are concatenated. 
		% the text is sensitive to some local changes like light and occlusion, and
	As the appearance and geometry features are able to capture different local information in the tracking process, combining them together is more robust. % than using one alone in video text tracking.
		%Although the text changes slowly in the global-level, it is easily affected locally by light changes, occlusion and so on. And the appearance and geometry features are able to capture different local information in the tracking process respectively
		%So combining them together is more robust than using one alone. 
	%Then, %the effect of different matching methods on performance has also been studied. 
	Then, in order to evaluate the effectiveness of the GRU in our proposed estimated descriptor, we try to associate $AGD$ of the adjacent frames directly, instead of matching the current frame $AGD$ with $EAGD$ obtained from the previous frame. Consequently, this matching method results in nearly 6\% loss on MOTA. This highlights that the temporal changing information captured by the GRU also plays an important role in the video text tracking.
	
	\subsection{Comparison with State-of-the-Art Video Text Detection Methods}
	\renewcommand{\arraystretch}{1.5} 
	\begin{table*}[tp]
		\centering
		\fontsize{9}{8}\selectfont
		\begin{threeparttable}
			\begin{tabular}{ccccccccccc}
				\toprule
				\multirow{2}{*}{Method}&
				\multicolumn{3}{c}{ ICDAR 2013}&\multicolumn{3}{c}{ Minetto dataset}&\multicolumn{3}{c}{ YVT} \cr
				\cmidrule(lr){2-4} \cmidrule(lr){5-7} \cmidrule(lr){8-10}
				&P&R&F&P&R&F&P&R&F&\cr
				\midrule
				Epshtein et al.~\cite{epshtein2010detecting} &39.80&32.53&35.94&$-$&$-$&$-$&68.00&{\bf76.00}&72.00 \cr
				Zhao et al.~\cite{zhao2011text}&47.02&46.30&46.65&$-$&$-$&$-$&34.00&41.00&37.00  \cr
				Minetto et al.~\cite{minetto2011snoopertrack}&$-$&$-$&$-$&61.00&69.00&63.00&$-$&$-$&$-$ \cr
				Yin et al.~\cite{yin2013robust}&48.62&54.73&51.56&$-$&$-$&$-$&$-$&$-$&$-$  \cr
				Moslen et al.~\cite{mosleh2013automatic}&$-$&$-$&$-$&$-$&$-$&$-$&79.00&72.00&75.00  \cr
				Wu et al.~\cite{wu2015new}&63.00&{\bf 68.00}&65.00&$-$&$-$&$-$&81.00&73.00&77.00 \cr
				Zuo et al.~\cite{zuo2015multi}&$-$&$-$&$-$&84.00&68.00&75.00&$-$&$-$&$-$   \cr
				Khare et al.~\cite{khare2017arbitrarily}&57.91&55.90&51.70&$-$&$-$&$-$&$-$&$-$&$-$  \cr
				Shivakumara et al.~\cite{shivakumara2017fractals}&61.00&57.00&59.00&$-$&$-$&$-$&79.00&73.00&76.00  \cr
				Pei et al.~\cite{pei2018scene}&$-$&$-$&$-$&89.00&84.00&86.00&$-$&$-$&$-$ \cr
				Wang et al.~\cite{wang2018scene}&58.34&51.74&54.45&88.80&87.53&88.14&$-$&$-$&$-$  \cr
				Wang et al.~\cite{wang2018robust} &71.90&58.67&62.65&83.03&84.22&83.30&$-$&$-$&$-$ \cr
				% Yang et al.~\cite{yang2018online}&$-$&$-$&$-$&91.00&87.00&89.00&$-$&$-$&$-$  \cr
				\midrule
				Our two-stage detection &79.97&55.21&65.32&82.90&80.66&81.77&70.15&64.52&67.22 \cr
				Our two-stage detection with online tracking &81.36&55.41&65.93&83.92&79.89&81.85&71.64&63.98&67.59 \cr
				Our end-to-end detection &80.10&57.07&66.65&86.69&{\bf91.17}&88.87&78.51&71.54&74.86 \cr
				Our end-to-end detection with online tracking &{\bf 82.36}&56.36&{\bf 66.92}&{\bf 91.27}&89.38&{\bf 90.32}&{\bf 89.12}&71.03&{\bf 79.05} \cr
				\bottomrule
			\end{tabular}
			\caption{\label{tab:detection}Comparison of video text detection  performances on several pubic datasets.}
		\end{threeparttable}
		\vspace{-0.95cm} 
	\end{table*}
	In this section, we compare our method with state-of-the-art methods on three public video text datasets. %In the results of each dataset, the detection result is on the top and the tracking result is at the bottom. For detection result, the deeper the color, the lower confidence of the bounding box, while boxes with the same color in different frames belong to the same trajectory for tracking result.
	As shown in Tab. \ref{tab:detection}, we have summarized various video text detection methods, and our method outperforms all the other methods by more than 2\% on f-measure. 
	
	The precision of video text detection is improved by a large margin along with high recall ratio with our method. This is mainly due to our end-to-end joint training and the introduction of long-term memory mechanisms. Specifically, in our end-to-end inference process, the detection results can be refined along with the updating of trajectories, and the long time memory can also inhibit some negative samples, resulting in more accurate results denoted by ``Our end-to-end detection with online tracking'' in Tab. \ref{tab:detection}.
	
	For training, except for the configuration introduced in Sec.~\ref{sec:id}, the YVT model is finetuned on its own training set as it contains many web videos. For testing, the longer sides of input images in ICDAR 2013 are resized to 1280. While, input images not resized for Minetto and YVT during evaluation. Besides, multi-scale testing is not conducted since it is too slow and unpractical for video especially.
	
	\section{Conclusions and Future Work}
	In this work, we present an end-to-end framework for video text detection with online tracking according to the characteristics of video scene text. The proposed $AGD$ and $EAGD$ are employed to convert the long-term multiple targets changing process to a trainable model. By sharing convolutional features, the text tracking branch is nearly cost-free. In the inference phase, the text detection results are obtained along with the trajectory online generation. Experiments on video text datasets show that our method significantly outperforms previous methods in both detection and tracking. However, there are still some drawbacks: the trajectory generation is not incorporated into the training process and semantic information has not been exploited. In the future, we plan to add video text detection, tracking, and recognition to an end-to-end framework.
	
	\section*{ACKNOWLEDGMENTS}
    This work is jointly supported by National Key Research
    and Development Program of China (2016YFB1001000),
    National Natural Science Foundation of China (61525306,
    61633021, 61721004, 61420106015, 61806194), Capital Science and Technology
    Leading Talent Training Project (Z181100006318030), 
    Beijing Science and Technology Project (Z181100008918010) and CAS-AIR.

	{\small
		\bibliographystyle{ieee}
		\bibliography{egbib}
	}
	
\end{document}